\documentclass[journal]{IEEEtran}

\usepackage[american]{babel}

\usepackage{mathtools}
\usepackage{amsmath,amssymb}
\usepackage{multirow}
\usepackage[caption=false]{subfig}
\usepackage{bm}
\usepackage{algorithm,algorithmic}
\usepackage{pifont}

\usepackage{epsfig}
\usepackage{graphicx}
\usepackage{booktabs}       
\usepackage{amsfonts}       
\usepackage{microtype}      
\usepackage{multirow}
\usepackage{color}
\usepackage{subfig}
\graphicspath{{figure/}}

\usepackage[breaklinks=true,colorlinks,bookmarks=false]{hyperref}

\begin{document}
\title{Piecewise classifier mappings: Learning fine-grained learners for novel categories with few examples}

\author{Xiu-Shen Wei,~\IEEEmembership{Member,~IEEE}, Peng Wang, Lingqiao Liu, Chunhua Shen,~\IEEEmembership{Member,~IEEE}\\Jianxin Wu,~\IEEEmembership{Member,~IEEE}
\thanks{$\bullet$ The first two authors contributed equally to this work. X.-S. Wei is with Megvii Research Nanjing, Megvii Technology, China. P. Wang is with University of Wollongong, Australia. L. Liu and C. Shen are with the University of Adelaide, Australia. J. Wu is with the National Key Laboratory for Novel Software Technology, Nanjing University, China.

$\bullet$ This work is supported by National Natural Science Foundation of China (61772256), DE170101259, and the program A for Outstanding Ph.D. candidate of Nanjing University (201702A010). C. Shen's participation was in part supported by the CRC GeoVision Project.

$\bullet$ Email: \{weixs.gm, wujx2001\}@gmail.com, \{lingqiao.liu, chunhua.shen\}@adelaide.edu.au, p.wang6@hotmail.com.}
}

\markboth{ACCEPTED BY IEEE TIP}%
{Shell \MakeLowercase{\textit{et al.}}: Bare Demo of IEEEtran.cls for IEEE Journals}

\maketitle

\begin{abstract}
Humans are capable of learning a new fine-grained concept with very little supervision, \emph{e.g.}, few exemplary images for a species of bird, yet our best deep learning systems need hundreds or thousands of labeled examples. In this paper, we try to reduce this gap by studying the fine-grained image recognition problem in a challenging few-shot learning setting, termed few-shot fine-grained recognition (FSFG). The task of FSFG requires the learning systems to build classifiers for novel fine-grained categories from few examples (only one or less than five). To solve this problem, we propose an end-to-end trainable deep network which is inspired by the state-of-the-art fine-grained recognition model and is tailored for the FSFG task.

Specifically, our network consists of a bilinear feature learning module and a classifier mapping module: while the former encodes the discriminative information of an exemplar image into a feature vector, the latter maps the intermediate feature into the decision boundary of the novel category. The key novelty of our model is a ``piecewise mappings'' function in the classifier mapping module, which generates the decision boundary via learning a set of more attainable sub-classifiers in a more parameter-economic way. We learn the exemplar-to-classifier mapping based on an auxiliary dataset in a meta-learning fashion, which is expected to be able to generalize to novel categories. By conducting comprehensive experiments on three fine-grained datasets, we demonstrate that the proposed method achieves superior performance over the competing baselines.
\end{abstract}

\section{Introduction}

Fine-grained recognition tasks such as identifying the species of birds~\cite{WahCUB200_2011}, dogs~\cite{Khosla11stanforddogs} and cars~\cite{cars}, have been popular in applications of computer vision. Since the categories are all similar to each other, different categories can only be distinguished by slight and subtle differences, which makes fine-grained recognition a challenging problem. Over the past decade, fine-grained recognition has attracted tremendous attention and observed rapid performance boost thanks to the integration of the sophisticated deep network structures with large annotated training datasets~\cite{Taomei17CVPR, Shao16CVPR, Max15NIPS, Di15CVPR, Tsungyu15ICCV, Yu16TIP, SCDATIP}.

However, the large-scale fine-grained data volume required to train such classification algorithms limits the ranges where they can be successfully applied to, \emph{e.g.}, very sparse training samples can be collected for some rare bird species. Humans, in contrast, are capable of learning a new fine-grained concept with very little supervision. To mimic this human ability, in this work, we study the fine-grained image recognition in a more practical and challenging few-shot setting, that is, we aim to learn the classifiers of novel fine-grained categories from very few labeled training examples (\emph{a.k.a.} exemplars, usually $1$ or $5$). 

\begin{figure}[t!]
\centering
	{\includegraphics[width=\columnwidth]{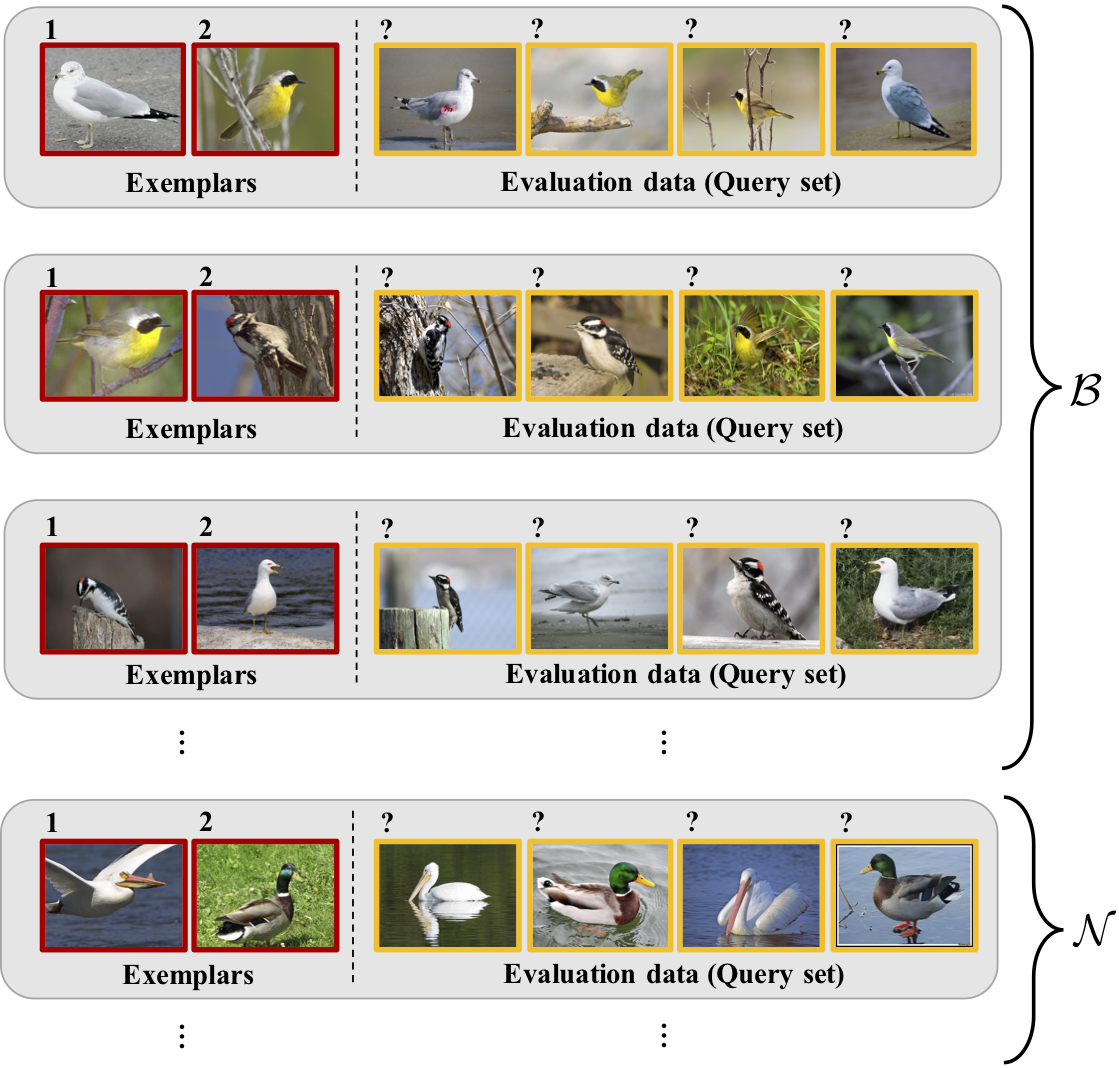}}
\vspace{-2em}
\caption{Illustration of the few-shot fine-grained image recognition (FSFG) task. The aim is to learn the classifier for a fine-grained category, bird species in this example, from few exemplars. We train the exemplar-to-classifier mapping based on an auxiliary dataset $\mathcal{B}$ and test the FSFG performance on another dataset $\mathcal{N}$. There are no category overlaps between these two sets.
}
\label{fig:fewshot}
\end{figure}

Learning a classifier for a fine-grained category identified by few exemplars is a challenging problem, as satisfactory classification performance can be expected only when the learned classifiers can capture the subtle differences between categories and is able to generalize beyond the very limited supervisions. To realize such exemplar-to-classifier mapping, we propose an end-to-end trainable network which is inspired by state-of-the-art fine-grained recognition model and is tailored for the FSFG task. Specifically, the network consists of a bilinear feature learning module and a classifier mapping module. While the former encodes the discriminative information of exemplar image into a feature vector, the latter, as the key part of the network, maps the intermediate image features into the category-level decision boundaries. Two problems remain to succeed with such mappings. On one hand, the distribution of the image-level representation can be complex which poses a great challenge for the mapping. On the other hand, the feature generated from bilinear pooling is very high dimensional, which further impedes the mapping due to the risk of parameter explosion.

The key novelty of our model to mitigate these problems is a ``piecewise mappings'' function in the classifier mapping module, which generates the decision boundary via learning a set of more attainable sub-classifiers in a much more parameter-economic way. Due to the outer product computation in bilinear pooling, the feature obtained, by nature, can be viewed as a set of sub-vectors, each of which implicitly attends to part of the image. We perform the sub-vector to sub-classifier mapping resorting to highly non-linear mappings. Then, these sub-classifiers are recombined into a global classifier so that it can tell samples from different categories. Intuitively, we learn the feature-to-classifier mapping based on the implicit ``part'' which may encode simpler and purer information and consequently makes the mapping easier. As a by-product, the piecewise mappings significantly reduce the number of model parameters and enable a more efficient computation. We learn the exemplar-to-classifier mapping using an auxiliary dataset in a meta-learning fashion as shown in Fig.~\ref{fig:fewshot}. The aim in the meta-training phase is to learn a ``mapping paradigm'' which is expected to be able to generalize to novel categories.

In experiments, we perform the proposed FSFG method on three fine-grained benchmark datasets, \emph{i.e.}, \emph{CUB Birds}~\cite{WahCUB200_2011}, \emph{Stanford Dogs}~\cite{Khosla11stanforddogs}, \emph{Stanford Cars}~\cite{cars}. Empirical results show that our FSFG model significantly outperforms competing baseline methods, including exemplar SVM~\cite{exemplarSVM11ICCV}, $k$-nearest neighbor and state-of-the-art generic few-shot learning methods~\cite{siamese16ICML, prototypical17NIPS,relation18CVPR}. Furthermore, we also conduct extensive ablation studies about our proposed method. These results could validate the effectiveness and efficiency of our FSFG model.

In summary, our major contributions are three-fold:
\begin{itemize}
\item We study the fine-grained image recognition in a challenging few-shot setting and propose a novel meta-learning strategy to address the FSFG problem.
\item We devise a novel exemplar-to-classifier mapping strategy, named piecewise mappings, which resorts to the special structure of the bilinear CNN features to learn a discriminative classifier in a parameter-economic way.
\item We conduct comprehensive experiments on three fine-grained benchmark datasets, and our proposed model achieves superior performance over competing solutions on all these datasets.
\end{itemize}


\section{Related work}\label{sec:related}

In this section, we briefly review the related work of both fine-grained image recognition and generic few-shot learning.

\subsection{Fine-grained image recognition}

Fine-grained recognition is a challenging problem and has recently emerged as an active topic~\cite{Khosla11stanforddogs, cars, WahCUB200_2011}. Over the past decade, fine-grained recognition has achieved high performance levels thanks to the integration of powerful deep learning techniques with large annotated training datasets. A number of effective fine-grained recognition methods have been developed in the literature~\cite{Steve14BMVC, Taomei17CVPR, Shao16CVPR, Max15NIPS, Di15CVPR, Tsungyu15ICCV, Yu16TIP}. Among them, some work, \emph{e.g.},~\cite{Max15NIPS, Tsungyu15ICCV}, attempted to learn a more discriminative feature representation by developing powerful deep models. Some methods aligned the objects in fine-grained images to eliminate pose variations and the influence of camera position, \emph{e.g.},~\cite{Steve14BMVC, Di15CVPR}. Moreover, some of them relied on localizing discriminative parts with/without strong supervisions, \emph{e.g.},~\cite{Taomei17CVPR, Shao16CVPR, Di15CVPR}.

However, current fine-grained recognition systems assume a set of categories known \emph{a priori}, despite the obviously dynamic and open nature of the visual world~\cite{feedforward16NIPS,Wang16ECCV,L2Lobjdec15CVPRdd}. Compared with previous work, we are studying fine-grained image recognition in a challenging few-shot learning setting where the model is required to recognize novel fine-grained categories by only a few labeled images.

\subsection{Generic few-shot image recognition}

Nowadays, few-shot image recognition (\emph{a.k.a.} few-shot learning or low-shot learning)~\cite{feedforward16NIPS, matchnet16NIPS,siyuanqiao18CVPR,Hang18CVPR} has attracted more and more attentions in computer vision and pattern recognition. This line of research explores the possibility of endowing learning systems the ability of rapid learning for novel categories from a few examples. More specifically, these systems are able to learn new concepts on the fly, from few or even a single example as in one-shot learning. Few-shot image recognition is usually tackled by using generative models~\cite{science,rezende16ICML} or, in a discriminative setting, using ad-hoc solutions such as exemplar support vector machines~\cite{exemplarSVM11ICCV}. While recently, many methods solved it in a learning-to-learn formulation~\cite{prototypical17NIPS, modelago17ICML, unsupervised16NIPS, L2Lobjdec15CVPRdd, Wang16ECCV, Serena17CVPR,Sachin2017,DBLP:journals/corr/LiZCL17}.

Specifically, in recent years, Vinyals et al.~\cite{matchnet16NIPS} proposed Matching Networks, which uses an attention mechanism over a learned embedding of the labeled set of examples (the support set) to predict classes for the unlabeled points (the query set). It can be interpreted as a weighted nearest-neighbor classifier applied within an embedding space. Later, Snell et al.~\cite{prototypical17NIPS} developed Prototypical Networks for generic few-shot learning. \cite{prototypical17NIPS} further improved~\cite{matchnet16NIPS} by considering there exists an embedding in which points cluster around a single prototype representation for each class. It achieved better classification accuracy than \cite{matchnet16NIPS} in the few-shot learning setting. In~\cite{Wang16ECCV}, it learned a regression network that maps from small-sample model parameters (\emph{i.e.}, small-sample decision boundary) to large-sample model parameters (\emph{i.e.}, large-sample decision boundary). Meanwhile, the method of~\cite{Wang16ECCV} was also performed in a meta-learning fashion. Additionally, in~\cite{Sachin2017}, the authors reformulated the parameter update into an LSTM and achieved this via a meta-learner. To solve new learning tasks with few samples, the method in~\cite{modelago17ICML} designed a so called model-agnostic meta-learning scheme, the essential idea of which is to require the parameters to be able to perform well on new task via one or few gradient steps on this task. The method in~\cite{DBLP:journals/corr/LiZCL17} took a step further by updating the model parameters as well as the learning rate in a uniform meta-learning framework.

More similar to our works are~\cite{siyuanqiao18CVPR,Hang18CVPR}, which attempted to train parameter predictors for novel categories also from activations. However, the most differences between ours and~\cite{siyuanqiao18CVPR,Hang18CVPR} are two-fold. 1) Our novel categories classifiers are learnt in a meta-learning fashion, while, \cite{siyuanqiao18CVPR,Hang18CVPR} employ traditional learning strategy. 2) More importantly, our proposed method can leverage the bilinear structure of powerful image representations for fine-grained objects. Besides, we also develop a novel classifier learning paradigm, \emph{i.e.}, piecewise classifier mappings (a.k.a. sub-classifier mapping), which can not only prevent overfitting caused by high-dimensionality of bilinear, but also have a good motivation for the few-shot fine-grained recognition task. Experimental results validate our proposal and prove our learning strategy design.

Additionally, many previous few-shot image recognition studies all focused on generic images (\emph{e.g.}, images of the ImageNet~\cite{ILSVRC15} and CIFAR~\cite{cifar} datasets) or generic patterns (\emph{e.g.}, characters of the Omniglot~\cite{cogsci2011} dataset). In fact, some generic few-shot learning methods, \emph{e.g.},~\cite{Wang16ECCV} and \cite{unsupervised16NIPS}, did consider fine-grained recognition scenarios and evaluate on fine-grained datasets. However, compared with those tasks, we specifically consider a novel few-shot image recognition topic, \emph{i.e.}, few-shot fine-grained image recognition. The most different point of our topic from the generic few-shot image recognition is that, fine-grained recognition relies on more subtle image cues which makes it considerably more challenging. We demonstrate that the proposed model, especially our piecewise mappings component, can cater to the desire of capturing the subtle differences in a fine-grained scenario from limited training data, even one-shot.

\section{Learning few-shot fine-grained learners}\label{sec:model}

In this section, we firstly present our learning strategy for FSFG and introduce the relevant notations. Then, a detailed elaboration of various aspects of our method will be followed in the subsequent sections. 

\subsection{Learning strategy and notations}

\begin{figure}[t!]
\centering
	{\includegraphics[width=0.95\columnwidth]{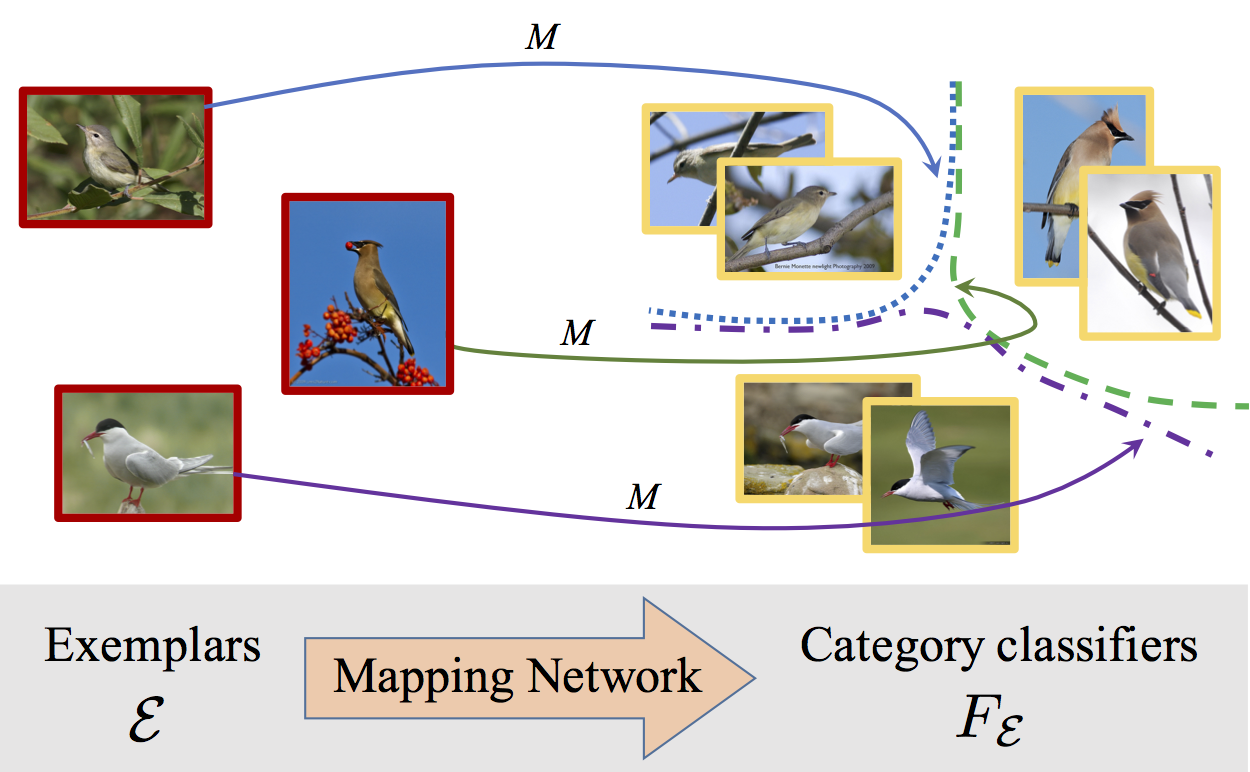}}
\caption{Key idea of the proposed FSFG model. In each episode, we sample an exemplar set $\mathcal{E}$ from $\mathcal{B}$, which is composed of a subset of categories (three categories in this example) and each category contains few exemplars (the images with red border). We wish to learn a mapping $M$ that can map these exemplars into their corresponding category classifiers (the dashed lines). The mapping parameters are learned so that these classifiers can correctly distinguish the query images (the images with yellow border).}
\label{fig:mapping}
\end{figure}

Our work is built upon the framework of meta-learning which treats the classifier generation process as a mapping function from the few labeled training samples of a category, called ``exemplars'' hereafter, to their corresponding category classifier. Fig.~\ref{fig:mapping} shows the key idea of this learning scheme. This \emph{exemplar-to-classifier} mapping is learned on an auxiliary training set $\mathcal{B}$. It contains $N$ labeled training images $\mathcal{B}=\{(\mathcal{I}_1,y_1),(\mathcal{I}_2,y_2),\ldots,(\mathcal{I}_N,y_N)\}$, where $\mathcal{I}_i$ is an example image and $y_i\in\{1,2,\ldots,C_{\mathcal{B}}\}$ is its corresponding label. Once the mapping function is learned, it will be applied on another testing set $\mathcal{N}$ to evaluate its performance, where $\mathcal{N}$ contains images of novel categories that do not appear in $\mathcal{B}$. 

To train the mapping function, we randomly sample a set of ``meta-training sets'' from $\mathcal{B}$. Each meta-training set (corresponding to a training episode) contains $C_{\mathcal{E}}<C_{\mathcal{B}}$ randomly chosen categories and a few images associated with them. A meta-training set is composed of an ``{exemplar set}'' $\mathcal{E}$ and a ``{query set}'' $\mathcal{Q}$ to mimic the scenario at the testing stage. Specifically, $\mathcal{E}$ contains $N_e$ (\emph{e.g.}, 1 or 5) exemplar images per category. The query set $\mathcal{Q}$ is coupled with $\mathcal{E}$ (has the same categories), but has no overlapped images. Each category of $\mathcal{Q}$ contains $N_q$ query images. During training, $\mathcal{E}$ will be fed into the to-be-learned mapping function $M$ to generate the category classifiers $F_{\mathcal{E}}$:
\begin{equation}\label{eq:mapping}
 \mathcal{E} \xrightarrow{M} F_{\mathcal{E}} \,.
\end{equation}
Then, $F_{\mathcal{E}}$ are subsequently applied to $\mathcal{Q}$ for evaluating the classification loss. The training objective then amounts to learning the mapping function by minimizing the classification loss. This process is formally written as follows:
\begin{equation}\label{eq:mapping}
 \min_{\lambda} \mathop{E}_{\{\mathcal{E},\mathcal{Q}\} \sim \mathcal{B}} \left\{ \mathcal{L}\left( F_{\mathcal{E}}\circ \mathcal{Q}\right) \right\},
\end{equation}where $\lambda$ denotes the model parameters of the mapping function $M$ (from $\mathcal{E}$ to $F_{\mathcal{E}}$), and $\mathcal{L}$ is the loss function.  $F_{\mathcal{E}}\circ \mathcal{Q}$ denotes applying the category classifiers $F_{\mathcal{E}}$ generated by the exemplar set $\mathcal{E}$ on the query set $\mathcal{Q}$.

\subsection{Model}

\begin{figure*}[t!]
\centering
	{\includegraphics[width=0.85\textwidth]{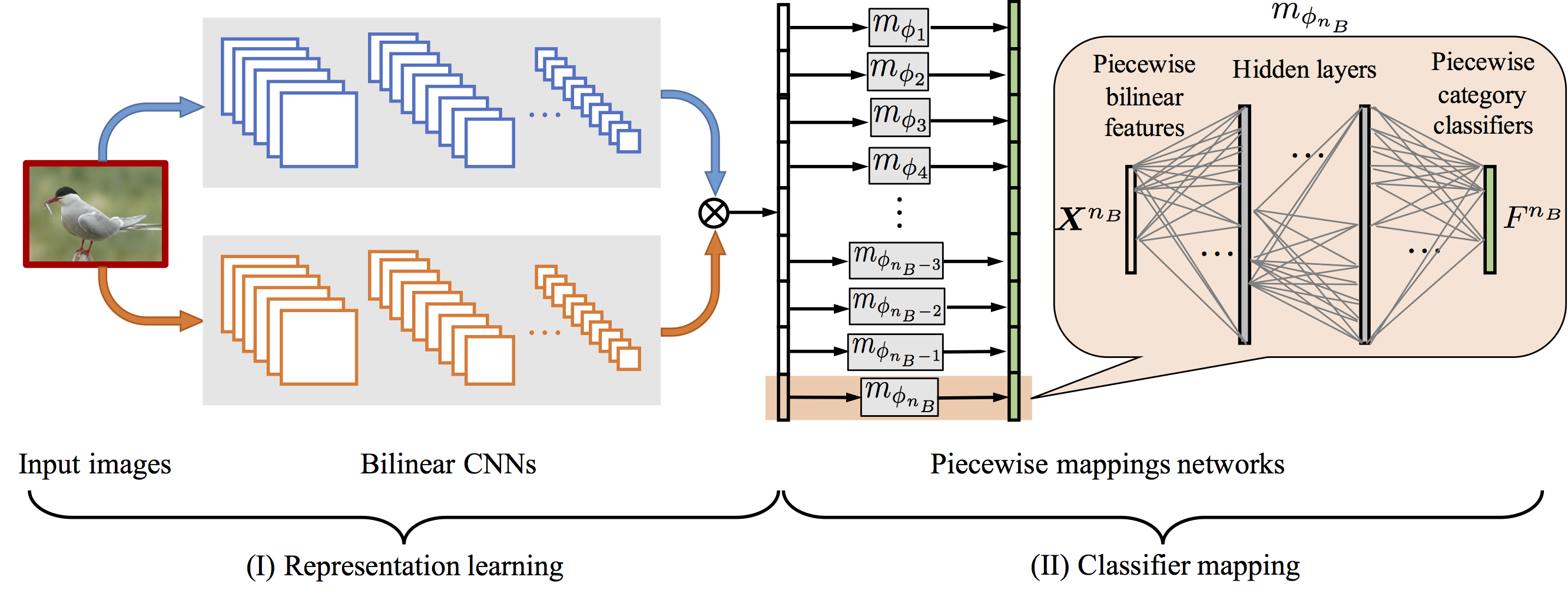}}
	\vspace{-1em}
\caption{Overview structure of our proposed FSFG model. On the left, it is the first component (the bilinear pooling module) for representation learning. On the right, the second component (the classifier mapping module) mapps the intermediate image features into the category classifiers.}
\label{fig:pipeline}
\end{figure*}

We implement the above exemplar-to-classifier mapping by adopting a trainable neural network. Fig.~\ref{fig:pipeline} shows the overall architecture of the network. As we can see, the network is composed of two modules: a representation learning module and a classifier mapping module. While the former adopts a bilinear CNN structure to encode the discriminative information of an exemplar image into a high-dimensional feature vector, the latter, as the key part of the network, maps the intermediate image representation into a category classifier. In the next two sub-sections, we elaborate these two modules in more details.

\subsubsection{Representation learning}\label{sec:representation}

\begin{figure}[t]
 \centering
 \subfloat[\emph{CUB Birds}]  { \includegraphics[width=0.8\columnwidth]{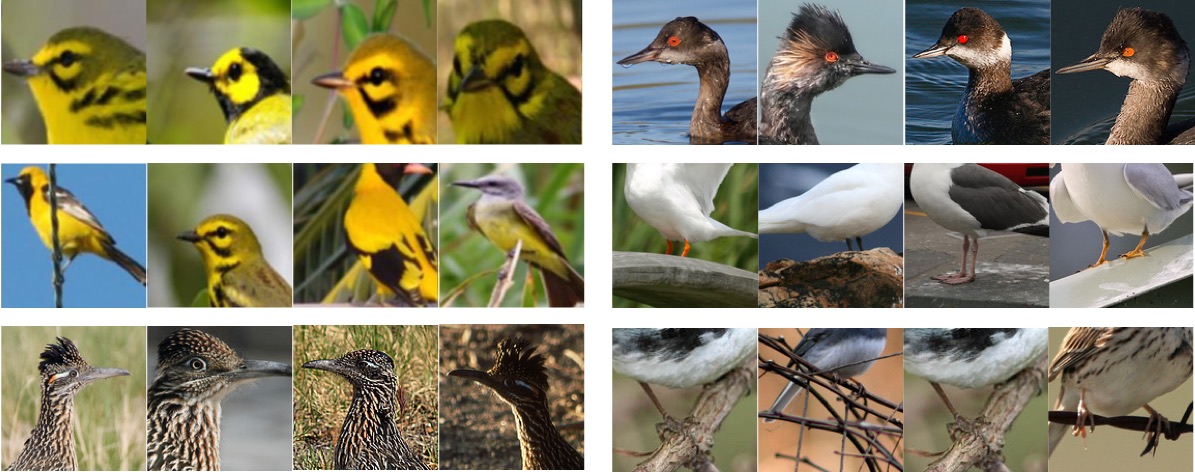} \label{fig:cubBCNN} }\\
 \subfloat[\emph{Stanford Dogs}] { \includegraphics[width=0.8\columnwidth]{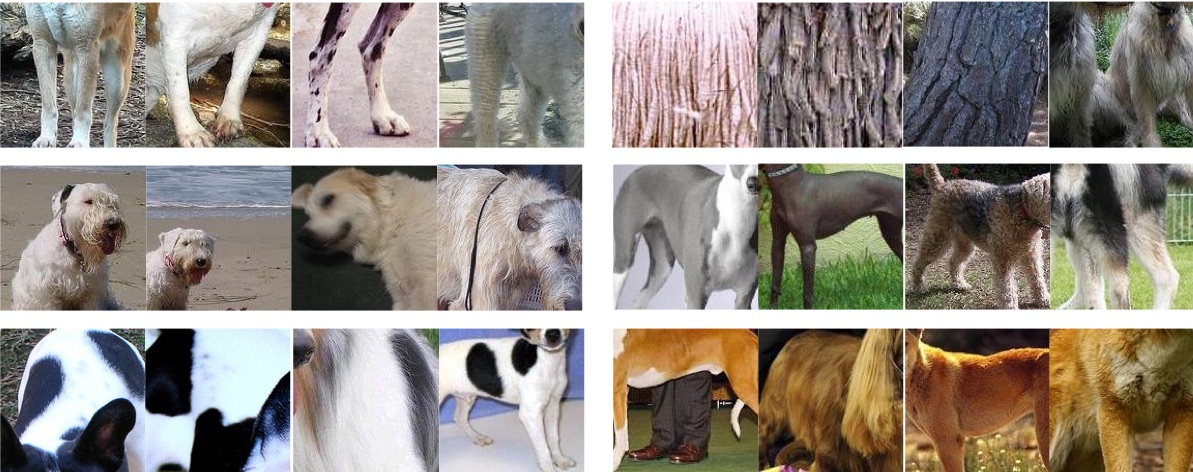} \label{fig:dogBCNN} }\\
 \subfloat[\emph{Stanford Cars}]  { \includegraphics[width=0.8\columnwidth]{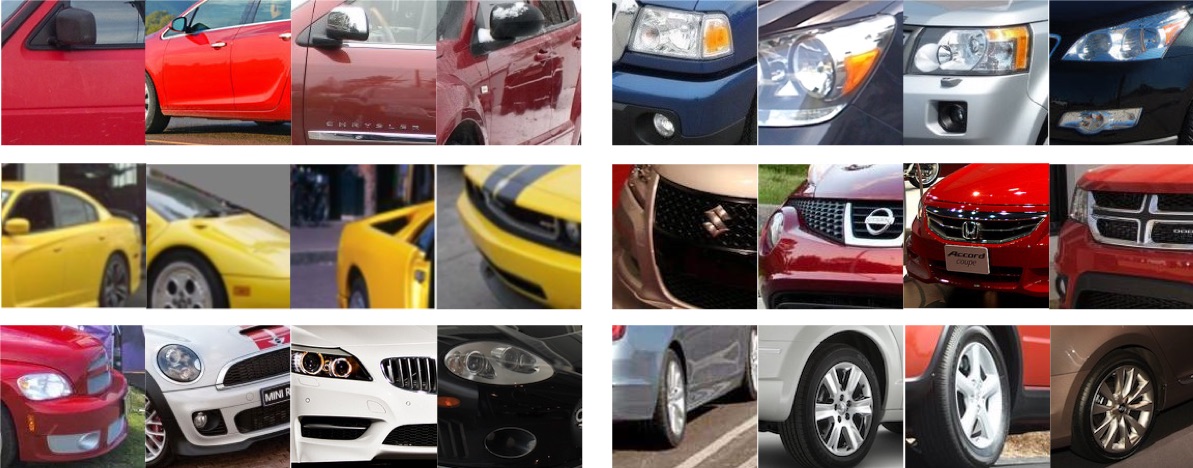} \label{fig:carBCNN} }
 \caption{``Parts'' of an object specifically correspond to the meaningful regions of the fine-grained objects, \emph{e.g.}, the beak of birds, the foot of dogs and the wheel of cars, etc. In the figures, by following~\cite{Tsungyu15ICCV}, we show the patches with the highest activation for several random filters of the BCNN models used in our experiments on three datasets, respectively.} \label{fig:bcnnpart}
\end{figure}

We employ a bilinear CNN (BCNN) structure \cite{Tsungyu15ICCV} to learn the image representation considering its state-of-the-art performance in fine-grained image recognition. BCNN consists of two feature extractors whose outputs are multiplied using outer product at each location of the image and pooled to obtain an image representation. Concretely, given two convolutional networks ($A$ and $B$) as two streams of BCNN, we assume their outputs are re-organized into $f_A(\mathcal{I})\in\mathbb{R}^{n_A\times{L}}$ and $f_B(\mathcal{I})\in\mathbb{R}^{n_B\times{L}}$, where $n_A$, $n_B$ denotes the dimensionality of the outputs and $L$ denotes the spatial locations. Then, at location $l$, the bilinear representation will be $\mathbf{b}_l\in\mathbb{R}^{n_A\times{n_B}}$,
\begin{equation}
\mathbf{b}_l=f_A(l,\mathcal{I}){f_B(l,\mathcal{I})}^{\top}\,.
\end{equation}
The vectorized versions of $\{\mathbf{b}_l\}$ will be pooled over the entire image to derive the image representation $\bm{x}\in\mathbb{R}^{D\times{1}}$ (for interpretation simplicity we let $D=n_A\times{n_B}$), that is,
\begin{equation}
\bm{x}(\mathcal{I}) = \sum_{l=1}^{L}{\rm{vec}}(\mathbf{b}_l)\,.
\end{equation}
With the outer product computation, bilinear structure modulates one feature stream with another. Thus, the BCNN feature $\bm{x}$ can be viewed as a set of $n_B$ sub-vectors $\bm{x}^t$:
\begin{equation}
\bm{x} = \left[ \bm{x}^1; \bm{x}^2; \ldots; \bm{x}^t; \ldots; \bm{x}^{n_B} \right] \,, \forall t: \bm{x}^t \in \mathbb{R}^{n_A\times 1} \,,
\end{equation}
where $\bm{x}^t$ is the modulated feature of $f_A$ by the $t$-th feature of $f_B$. This is similar to the multiplicative feature interactions in attention mechanisms~\cite{Tsungyu15ICCV}. From the observation that each modulated feature map tends to focus on an implicit ``part'' of an object (cf. Fig.~\ref{fig:bcnnpart}), and thus, $\bm{x}^t$ can be viewed as the feature description for that ``part''. In our implementation, we train the bilinear CNN by performing the same procedure in~\cite{Tsungyu15ICCV} and use it as the image representation extractor.

To represent a set of $N_e$ exemplar images belonging to category $k$, we simply compute the mean image representation as the category-level representation by:
\begin{equation}\label{eq:finalXe}
\bm{X}_k=\frac{1}{N_e}\sum_{i=1}^{N_e}\bm{x}_i \,,
\end{equation}
where $\{\bm{x}_i\}$ are samples with $y_i=k$.

\subsubsection{Classifier mapping}\label{sec:classifier_mapping}

Now that the information of each category identified by few exemplars has been encoded into a bilinear feature vector, the task of the classifier mapping module is to map these intermediate category-level representations into their corresponding category classifiers. Mathematically, this module computes a $D$-dimensional classifier $F_k\in\mathbb{R}^{D}$ for each category through a mapping $M:\mathbb{R}^{D}\rightarrow\mathbb{R}^{D}$. 

A straightforward solution to realize this mapping is via a global mapping, either linear or nonlinear. For example, a linear mapping can be:
\begin{equation}
F_k = \mathbf{W}_g\bm{X}_k +\mathbf{b}_g\,,
\end{equation}
where $\mathbf{W}_g\in\mathbb{R}^{D\times{D}}$ and $\mathbf{b}_g\in\mathbb{R}^{D}$ denote the parameters of the global mapping. However, this mapping strategy suffers from two drawbacks. First, as the feature $\bm{X}_k$ is supposed to encode the category-level information, the distribution of which can be highly complex. This poses a great challenge for the global mapping to find a decision boundary in such a complex feature space. Second, since the bilinear feature tends to be high dimensional, this mapping may result in parameter explosion, which will make the network training hard or infeasible. 

To mitigate these problems, we propose a novel ``piecewise mappings'' strategy, which exploits the structure of the bilinear features. As analyzed in Sec.~\ref{sec:representation}, the bilinear feature $\bm{X}_k$ can be viewed as a set of sub-vectors $\bm{X}^t_k$ with each sub-vector describes an implicit ``part'' of the object. Intuitively, we can test if an object falls into the category described in the exemplars by checking whether each ``part'' of it is compatible with the exemplars. This motivates us to apply a piecewise mapping to first map each sub-vector $\bm{X}^t_k$ into its corresponding sub-classifier $F_{k}^t$, and then combine these sub-classifiers together to generate the global category classifier. Fig.~\ref{fig:pipeline} shows this mapping with more details.

Concretely, a sub-vector $\bm{X}_k^t$ is firstly mapped into a sub-classifier $F_{k}^t$ via a nonlinear multilayer perceptron (MLP) $m_{\phi_t}(\cdot)$ as
\begin{equation}\label{eq:F}
F_{k}^{t} = m_{\phi_t}(\bm{X}_k^t)\,.
\end{equation}
We learn $n_B$ such MLPs $\{m_{\phi_t}(\cdot)\}$ to derive $n_B$ sub-classifiers $\{F_{k}^t\}$, and then these sub-classifiers are concatenated together to generate the global category classifier $F_k$:
\begin{equation}\label{eq:wholeF}
F_k = [F_{k}^1; F_{k}^2; \ldots; F_{k}^{n_B}]\,.
\end{equation}

Essentially, our model simplifies the global mapping approach by assuming that the classifier for the $t$-th sub-vector is solely determined by the information from the $t$-th sub-vector in the exemplar set. Despite resulting more restrictive mapping function, this assumption makes the network much easier to train.  
Note that, this mapping scheme will significantly reduce the model parameters involved in classifier generation. Taking one-layer mapping for example, let's assume $n_A=n_B=512$. For the global mapping, it requires more than $512^4$ parameters. For the proposed piecewise mappings, however, the number is reduced to about $512^3$. In addition, although there are parameter-economy variants of BCNN~\cite{CB16CVPR}, our piecewise classifier mappings still show better performance. This suggests that the proposed classifier mapping function brings benefits more than merely reducing the model size (cf. Table~\ref{table:result}).

\subsubsection{Network training}

Given a query sample $\bm{x}$ with label $y=c$, we compute its prediction distribution via softmax as:
\begin{equation}\label{eq:likelihood}
p_M(y=c|\bm{x})=\frac{\exp(F_c\cdot{\bm{x}})}{\sum_{c'}{\exp(F_{c'}\cdot{\bm{x}})}}\,.
\end{equation}
The model parameters are trained via minimizing the negative log-likelihood $\mathcal{J}(\bm{x},y) = -\log(p_M(c|\bm{x}))$. With this, we can now summarize the training in an episode as follows. First, we select an exemplar set $\mathcal{E}$ from $\mathcal{B}$ and learn/generate the classifiers $F_{\mathcal{E}}$. Then, we establish a query set $\mathcal{Q}$. The model parameters are optimized by minimizing $\mathcal{J}(\mathcal{Q})$. Algorithm~\ref{algo:episode} illustrates the training process in more details.
\begin{algorithm}[t]
\caption{Training episode loss computation for the proposed piecewise mappings.}
\label{algo:episode}
\small
\begin{algorithmic}[1]{
\REQUIRE {$\mathcal{B}$ is an auxiliary training set with $N$ images belonging to $C_{\mathcal{B}}$ categories; $\mathcal{B}_c$ denotes a subset of $\mathcal{B}$ containing all images belonging to the $c$-th category; $C_{\mathcal{E}}$ denotes the number of categories in an exemplar set $\mathcal{E}$ as well as a query set $\mathcal{Q}$ for an episode; $\mathcal{E}_k$ denotes the elements $(\bm{x}_i,y_i=k)$ in $\mathcal{E}$ with element size $N_e$; $\mathcal{Q}_k$ denotes the elements $(\bm{x}_j,y_j=k)$ in $\mathcal{Q}$ with element size $N_q$; $n$ denotes the number of piecewise mappings; \texttt{RandomSample}($\mathcal{T},N$) denotes a set of $N$ elements chosen uniformly at random from set $\mathcal{T}$, without replacement; $\mathcal{S}$ denotes a category set and ${S}_i$ denotes its $i$-th element.} 

\STATE {Select a category subset $\mathcal{S}$ for an episode\\ ${\mathcal{S} \leftarrow \rm \texttt{RandomSample}}(\{1,2,\ldots,C_{\mathcal{B}}\},C_{\mathcal{E}}$)};
\FOR{$k {\rm~in~} \{1,2,\ldots,C_{\mathcal{E}}\}$}
\STATE {Select  $\mathcal{E}_{k} \leftarrow {\rm \texttt{RandomSample}}(\mathcal{B}_{{S}_k},N_e$)};
\STATE {Compute the category-level representation $\bm{X}_{k}$ following Eq.~\ref{eq:finalXe}};
\STATE {Generate the category classifier $F_k$ by Eq.~\ref{eq:F}} and Eq.~\ref{eq:wholeF};
\STATE {Select $\mathcal{Q}_{k} \leftarrow {\rm \texttt{RandomSample}}(\mathcal{B}_{{S}_k}\backslash \mathcal{E}_{k},N_q$)};
\ENDFOR
\STATE {Initialize loss $\mathcal{J}\leftarrow 0$};

\FOR{$k {\rm~in~}  \{1,2,\ldots, C_{\mathcal{E}}\}$}
    \FOR{$(\bm{x},y)$ in $\mathcal{Q}_{k}$}
    		\STATE {$\mathcal{J}\leftarrow \mathcal{J}+\mathcal{J}(\bm{x},y)$};
    \ENDFOR   
\ENDFOR
\STATE{$\mathcal{J}=\frac{\mathcal{J}}{C_{\mathcal{E}}\times{N_q}}$}	

\STATE {Update model parameters by minimizing $\mathcal{J}$};
\RETURN {$n$ piecewise mappings $\left[ m_{\phi_1}; \ldots; m_{\phi_n}\right]$}.
}\end{algorithmic}
\end{algorithm}

\section{Experiments}\label{sec:experiment}

In this section, we first describe the experimental settings, and then present the main results. Later, Ablation studies are given to further evaluate the effectiveness of our proposed classifier mapping strategy.

\subsection{Datasets, setups and implementation details}\label{sec:setting}

Our experiments are conducted on three fine-grained benchmark datasets, \emph{i.e.}, \emph{CUB Birds} ($200$ categories of birds, $11,788$ images)~\cite{WahCUB200_2011}, \emph{Stanford Dogs} ($120$ categories of dogs, $20,580$ images)~\cite{Khosla11stanforddogs}, \emph{Stanford Cars} ($196$ categories of cars, $16,185$ images)~\cite{cars}. For each dataset, we randomly split its original image categories into two disjoint subsets: one as the auxiliary training set $\mathcal{B}$, and the other as the FSFG testing set $\mathcal{N}$. Table~\ref{table:dataset} presents the details of the category split. For each category in $\mathcal{B}$, we follow the raw splits provided by these datasets to split the data into training and validation. While the former is used to train the parameters, the latter is used to monitor the learning process. 

\begin{table}[t]
 \caption{Category split for three datasets. ${C}_{\rm total}$ denotes the total number of categories in a dataset, $C_{\mathcal{B}}$ denotes the number of categories in $\mathcal{B}$  and $C_{\mathcal{N}}$ denotes the number of categories in $\mathcal{N}$.} \label{table:dataset}
 \centering
 \footnotesize
 \begin{tabular}{|c|c|c|c|}
  \hline
	$\sharp$ category & \emph{CUB Birds} & \emph{Stanford Dogs} & \emph{Stanford Cars} \\
  \hline
  \hline
	$C_{\rm total}$ & 200 & 120 & 196 \\
	$C_{\mathcal{B}}$ & 150 & 90 & 147 \\
	$C_{\mathcal{N}}$ & 50 & 30 & 49 \\
  \hline
 \end{tabular}
\end{table}

To mimic the testing condition, in each training episode, we set the category size of the exemplar set $\mathcal{E}$ to be same as the number of categories in the testing set $\mathcal{N}$, \emph{i.e.}, $C_{\mathcal{E}} = C_{\mathcal{N}}$. Further we set $N_e=1$ ($N_e=5$) for one-shot learning (five-shot learning) and $N_q$ is set to be $20$ in all settings (by following the protocol in~\cite{prototypical17NIPS}). Similarly, during the testing phase, for each category in $\mathcal{N}$, we randomly choose one exemplar (five exemplars) for one-shot learning (five-shot learning), and another $20$ samples are randomly selected to evaluate the recognition performance. We repeat this evaluation process twenty times, and the mean classification accuracy is used as the evaluation criterion.

In theory, we can choose any network structures as the base network for our bilinear feature learning module. Since our key contribution is in the classifier mapping scheme, we choose AlexNet~\cite{cnn12} as the two streams in BCNN, considering the trade off between its representation capacity and computational efficiency. Specifically, we adopt the AlexNet model pre-trained on the Places 205 database~\cite{placescnn} to initialize the representation learning parameters. The reason why we use the Place dataset~\cite{placescnn} instead of ImageNet~\cite{ILSVRC15} is to avoid the FGFS testing categories to be present in the pre-training dataset. We fine-tune the bilinear feature learning module on the auxiliary training set first and freeze it during the classifier learning process. For the classifier mapping module, without otherwise stated, we choose the mapping function $m_{\phi_t}$ to be a three-layer MLP, where $1024$ hidden units are adopted in each layer and Exponential Linear Units (ELU)~\cite{elu} is used in each layer as the non-linear activation function. SGD is used to optimize the parameters with learning rate of $0.1$. We implement our model using the open-source library PyTorch.

\subsection{Main results}

We present the main results of FSFG by firstly introducing some baseline methods and then reporting the empirical results on these three datasets.

\subsubsection{Comparison methods}

\begin{table*}[t]
 \caption{Comparison results (mean$\pm$std.) on three fine-grained datasets. The highest average accuracy of each column is marked in bold. ``$\bullet$/$\circ$'' denotes that our proposed model performs significantly better/worse than the corresponding method by the pairwise $t$-test with confidence level 0.05. ``FB'' stands for using the fully bilinear pooling representations, and ``CB'' is for using compact bilinear pooling.} \label{table:result}
 \centering
\footnotesize
 \begin{tabular}{|c|c|c|c|c|c|c|}
  \hline
  \multirow{2}{*}{Method} & \multicolumn{2}{c|}{\textit{CUB Birds}} & \multicolumn{2}{c|}{\textit{Stanford Dogs}} & \multicolumn{2}{c|}{\textit{Stanford Cars}} \\
  \cline{2-7} & 1-shot & 5-shot & 1-shot & 5-shot & 1-shot & 5-shot  \\
  \hline
  \hline
  $k$-NN (FB) &  38.85{\scriptsize$\pm$3.43} $\bullet$  &  55.58{\scriptsize$\pm$0.84} $\bullet$   &  24.53{\scriptsize$\pm$2.36} $\bullet$   &  40.30{\scriptsize$\pm$2.34} $\bullet$   &  26.99{\scriptsize$\pm$2.91} $\bullet$  &  43.40{\scriptsize$\pm$1.68} $\bullet$   \\
  $k$-NN (CB) &  24.52{\scriptsize$\pm$1.80} $\bullet$ & 41.85{\scriptsize$\pm$1.51} $\bullet$ &  18.31{\scriptsize$\pm$1.81} $\bullet$ & 32.37{\scriptsize$\pm$1.15} $\bullet$   & 21.25{\scriptsize$\pm$1.78} $\bullet$   &   39.42{\scriptsize$\pm$1.57} $\bullet$  \\
  SVM (FB) &  34.47{\scriptsize$\pm$1.93} $\bullet$  &  59.19{\scriptsize$\pm$1.28} $\bullet$  &  23.37{\scriptsize$\pm$3.18} $\bullet$  &  39.50{\scriptsize$\pm$1.07} $\bullet$  & 25.66{\scriptsize$\pm$1.53} $\bullet$   &  51.07{\scriptsize$\pm$1.51}~~~   \\
  SVM (CB) & 24.94{\scriptsize$\pm$1.97} $\bullet$   & 41.93{\scriptsize$\pm$1.69} $\bullet$   &  18.25{\scriptsize$\pm$2.83} $\bullet$  &  30.50{\scriptsize$\pm$1.76} $\bullet$  &  21.34{\scriptsize$\pm$1.94} $\bullet$  &   39.43{\scriptsize$\pm$1.46} $\bullet$  \\
  Siamese-Net (FB)~\cite{siamese16ICML} &    37.38{\scriptsize$\pm$1.53} $\bullet$  &  57.73{\scriptsize$\pm$1.38} $\bullet$  &  23.99{\scriptsize$\pm$1.66} $\bullet$  &  39.69{\scriptsize$\pm$1.17} $\bullet$  & 25.81{\scriptsize$\pm$1.67} $\bullet$   &  48.95{\scriptsize$\pm$1.31} $\bullet$  \\
  Siamese-Net (CB)~\cite{siamese16ICML} &    26.58{\scriptsize$\pm$2.47} $\bullet$  &  43.51{\scriptsize$\pm$1.53} $\bullet$  &  19.28{\scriptsize$\pm$2.60} $\bullet$  &  31.49{\scriptsize$\pm$1.22} $\bullet$  & 22.41{\scriptsize$\pm$1.55} $\bullet$   &  40.07{\scriptsize$\pm$1.88} $\bullet$   \\
  Prototypical Network (FB)~\cite{prototypical17NIPS} &    38.96{\scriptsize$\pm$1.43} $\bullet$  &  58.62{\scriptsize$\pm$1.65} $\bullet$  &  25.05{\scriptsize$\pm$1.34} $\bullet$  &  40.42{\scriptsize$\pm$1.54} $\bullet$  & 25.33{\scriptsize$\pm$1.87} $\bullet$   &  49.03{\scriptsize$\pm$1.60} $\bullet$  \\
  Prototypical Network (CB)~\cite{prototypical17NIPS} &    28.88{\scriptsize$\pm$1.41} $\bullet$  &  44.28{\scriptsize$\pm$1.57} $\bullet$  &  21.40{\scriptsize$\pm$1.24} $\bullet$  &  32.99{\scriptsize$\pm$2.11} $\bullet$  & 24.48{\scriptsize$\pm$1.67} $\bullet$   &  42.91{\scriptsize$\pm$1.18} $\bullet$   \\
 Relation Network (FB)~\cite{relation18CVPR} &    {39.68\scriptsize$\pm$1.19} $\bullet$  &  {59.39\scriptsize$\pm$1.50} $\bullet$  &  {26.11\scriptsize$\pm$1.14} $\bullet$  &  {41.55\scriptsize$\pm$1.88} $\bullet$  & {25.98\scriptsize$\pm$1.30} $\bullet$   &  {49.66\scriptsize$\pm$1.19} $\bullet$  \\
  Relation Network (CB)~\cite{relation18CVPR} &    {30.01\scriptsize$\pm$1.11} $\bullet$  &  {45.19\scriptsize$\pm$1.25} $\bullet$  &  {22.96\scriptsize$\pm$1.58} $\bullet$  &  {33.81\scriptsize$\pm$1.69} $\bullet$  & {25.74\scriptsize$\pm$1.77} $\bullet$   &  {44.09\scriptsize$\pm$1.53} $\bullet$   \\
  Global mapping (FB-) &    24.12{\scriptsize$\pm$1.39} $\bullet$  &   34.59{\scriptsize$\pm$1.77} $\bullet$ &  20.55{\scriptsize$\pm$1.48} $\bullet$  &  30.93{\scriptsize$\pm$1.91} $\bullet$  &  20.50{\scriptsize$\pm$1.60} $\bullet$  &   30.58{\scriptsize$\pm$1.82} $\bullet$  \\
  Global mapping (CB) &  25.42{\scriptsize$\pm$2.22} $\bullet$  &   36.37{\scriptsize$\pm$1.04} $\bullet$ &  20.77{\scriptsize$\pm$2.75} $\bullet$  &  32.33{\scriptsize$\pm$2.11} $\bullet$  &  20.24{\scriptsize$\pm$1.94} $\bullet$  &   32.66{\scriptsize$\pm$1.86} $\bullet$  \\
  \hline
  \hline
  Ours & \textbf{42.10{\scriptsize$\pm$1.96}}~~~    &  \textbf{62.48{\scriptsize$\pm$1.21}}~~~  & \textbf{28.78{\scriptsize$\pm$2.33}}~~~   &  \textbf{46.92{\scriptsize$\pm$2.00}}~~~  &  \textbf{29.63{\scriptsize$\pm$2.38}}~~~  &   \textbf{52.28{\scriptsize$\pm$1.46}}~~~  \\
  \hline
 \end{tabular}
\end{table*}

In our experiments, we compare our proposed model to the following competitive baselines. Apart from the original bilinear CNN, we also implement a compact bilinear CNN~\cite{CB16CVPR} as the image feature extractor to facilitate the comparison, which enables much lower feature dimensionality but keeps almost the same classification discriminative ability~\cite{CB16CVPR}. For compact bilinear pooling, we follow the optimal settings suggested in~\cite{CB16CVPR}. The dimensionality of compact bilinear pooling representations is $8,192$-d (much less than $65,536$-d of fully bilinear pooling). In our empirical results, the results of compact bilinear pooling are denoted as ``CB'' in Table~\ref{table:result}, and the results of fully bilinear pooling are denoted as ``FB''. Note that, most existing methods for generic few-shot learning are not applicable to our problem due to the formidable computation cost on high-dimensional bilinear features.

\begin{itemize}
\item \textbf{$k$-NN} ($k$-nearest neighbors): Following the testing setting introduced in Sec.~\ref{sec:setting}, we choose one sample (five samples) for each category in $\mathcal{N}$ as exemplar(s) and $20$ samples in the same category for evaluation. We use the BCNN (either original or compact version) fine-tuned on $\mathcal{B}$ as the image representation extractor, and nearest neighbor is adopted as the classifier to categorize the evaluation images. Specifically, the image representations are first $\ell_2$-normalized and cosine distance is used as the distance metric. Note that, for five-shot learning, the representations of five exemplars are averaged before normalization to serve as the category-level representation. This process will be repeated twenty times as for our method. (This applies to all other baselines, so we omit this when introducing the following baselines.)

\item \textbf{SVM} (support vector machine): After obtaining the bilinear representations for exemplars of the testing categories in $\mathcal{N}$, we train a classifier for each category based on these representations. In particular, for one-shot learning, this baseline becomes exemplar-SVMs~\cite{exemplarSVM11ICCV}.

\item \textbf{Siamese-Net}~\cite{siamese16ICML}: As a standard metric-learning strategy, Siamese-Net is a competitive solution for few-shot learning. It learns a feature space in which images of the same category are close but images belonging to different categories are separated apart. We train a Siamese-Net based on $\mathcal{B}$ by sampling pair-wise examples and the corresponding binary labels (``1'' presents examples are from the same category and ``0'' is not.) Similar to~\cite{siamese16ICML}, the regularized cross-entropy loss on the binary classifier is used. During evaluation, Siamese-Net could rank similarities between exemplars and testing data.

\item \textbf{Prototypical Network}~\cite{prototypical17NIPS} is one of state-of-the-art generic few-shot learning methods. It learns a metric space via the meta-learning fashion. In the learned metric space, classification can be performed by computing distances to prototype representations of each class. Here, we compare it as a strong baseline in our few-shot fine-grained setting.

\item \textbf{Relation Network}~\cite{relation18CVPR} is recently proposed for dealing with the few-shot generic image recognition problem. It develops a novel meta learning paradigm for few-shot learning. Specifically, a Relation Network is able to classify images of few classes by computing relation scores between query images and the few examples of each new class. Different from the other previous few-shot learning methods whose learning process occurs in the feature embedding, Relation Network can be seen as both learning a deep embedding and learning a deep non-linear metric (\emph{i.e.}, a similarity function).

\item \textbf{Global mapping}: As aforementioned in Sec.~\ref{sec:classifier_mapping}, an alternative solution to our proposed piecewise classifier mappings is global mapping. It follows the idea of the global feature to global classifier mapping by applying the mapping function directly on the category-level representation.
\end{itemize}

\subsubsection{Comparison results}

Table~\ref{table:result} presents the average accuracy rates of FSFG on the novel categories of three fine-grained datasets. For each dataset, we report both one-shot and five-shot recognition results. As shown in that table, our proposed model consistently and significantly outperforms the other baseline methods on these datasets.

\begin{table*}[t]
 \caption{Comparison results of global mapping and piecewise mappings (our proposal) on three datasets. The highest average accuracy of each column is marked in bold. ``$\bullet$'' denotes that the piecewise mappings outperform the global mapping with confidence level $0.05$ by the pairwise $t$-test.} \label{table:piecewise}
 \centering
 \footnotesize
 \begin{tabular}{|c|c|c|c|c|c|c|}
  \hline
  \multirow{2}{*}{Method} & \multicolumn{2}{c|}{\textit{CUB Birds}} & \multicolumn{2}{c|}{\textit{Stanford Dogs}} & \multicolumn{2}{c|}{\textit{Stanford Cars}} \\
  \cline{2-7} & 1-shot & 5-shot & 1-shot & 5-shot & 1-shot & 5-shot  \\
  \hline
  \hline
  Global mapping & 27.36{\scriptsize$\pm$1.64} $\bullet$   &  38.05{\scriptsize$\pm$1.55} $\bullet$  & 19.55{\scriptsize$\pm$2.27} $\bullet$   & 32.53{\scriptsize$\pm$2.35} $\bullet$   &  16.06{\scriptsize$\pm$2.06} $\bullet$  &  26.17{\scriptsize$\pm$1.02} $\bullet$   \\
  \hline
  Piecewise mappings (Ours) &  \textbf{31.00{\scriptsize$\pm$2.85}}~~~  &  \textbf{48.80{\scriptsize$\pm$2.33}}~~~  &   \textbf{23.07{\scriptsize$\pm$3.24}}~~~ &  \textbf{41.02{\scriptsize$\pm$2.50}}~~~  &  \textbf{18.98{\scriptsize$\pm$2.18}}~~~  &  \textbf{31.51{\scriptsize$\pm$1.38}}~~~   \\
  \hline
 \end{tabular}
\end{table*}

Generally, we see the simple baseline $k$-NN performs well and it even outperforms other more sophisticated baselines on some settings, \emph{e.g.}, on \emph{Stanford Dogs}. This is due to the discriminative capacity of the bilinear CNN features. SVM observes more obvious advantage comparing to $k$-NN when exploiting five training exemplars. Siamese-Net, as another discriminative method, achieves comparable performance to SVM but is outperformed by our method. In addition, our proposed method also outperforms Prototypical Networks and Relation Networks by a large margin. This reflects our meta-learning strategy can better generalize to unseen/novel fine-grained categories. For the global mapping, because BCNN generates image representation of ultra-high dimensionality (\emph{i.e.}, $65,536$ in our case), it is infeasible to learn a global mapping on such high-dimensional feature vectors. In order to realize the global mapping, we apply an additional linear mapping to first reduce $65,536$-d features into $8,192$-d feature vectors, and based on the low-dimensional features, we conduct the global mapping. It is denoted as ``Global mapping (FB-)'' in Table~\ref{table:result}. Specifically, the global mapping is also implemented as a three-layer networks. As seen, our proposed piecewise mappings significantly outperforms the global mapping. In ablation studies, we will further compare these two types of mapping schemes. 

Another interesting observation here is that the few-shot recognition performance gap between FB and CB is large. Note that, both FB and CB are trained on the same training set and achieve comparable classification performance on the validation set. This phenomenon may be explained as that the CB feature is not suitable for similarity matching (\emph{i.e.}, the experimental case of the testing set). It is an open problem worth future explorations.

In addition, we further investigate whether the proposed piecewise mapping idea works for existing generic few-shot learning approaches. Concretely, we apply our piecewise mapping module on the popular Prototypical Networks by learning a set of prototype features from the sub-vectors of the bilinear features. And the final classification of a sample is achieved by fusing the prediction scores from all the sub-features. Using similar hyper-parameters as the original Prototypical Networks, the modified Prototypical Networks achieve 40.16\%{\scriptsize$\pm$1.37\%} (60.18\%{\scriptsize$\pm$1.43\%}), 26.98\%{\scriptsize$\pm$1.32\%} (42.55\%{\scriptsize$\pm$1.65\%}), 27.41\%{\scriptsize$\pm$1.35\%} (51.49\%{\scriptsize$\pm$1.37\%}) recognition accuracy in the one-shot (five-shot) setting on \emph{CUB Birds}, \emph{Stanford Dogs} and \emph{Stanford Cars}, respectively. By comparing with Table~\ref{table:result}, the performance of the Prototypical Networks with piecewise mappings consistently outperforms original Prototypical Networks on all splits, which shows the effectiveness of our piecewise manner.



\subsection{Ablation studies}

To further inspect our piecewise mappings strategy for FSFG, we conduct ablation experiments on two aspects. First, we compare the global mapping and piecewise mappings on a fairer setting. Second, we investigate the influence of the mapping function $m_{\phi_{t}}$ variations on the FSFG performance. Finally, we also change the number of piecewise mappings (\emph{i.e.}, $n_B$) to show its stability.

\subsubsection{Piecewise mappings \emph{vs.} global mapping}\label{sec:PMvsGM}

\begin{figure}[t]
 \centering
 \subfloat[By global mapping]  { \includegraphics[width=0.48\columnwidth]{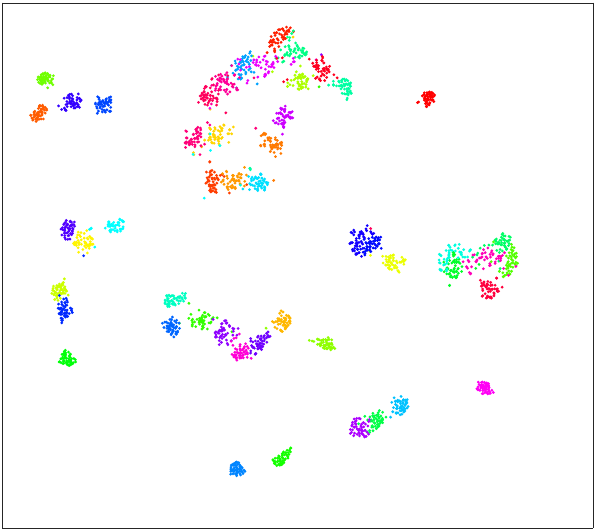} \label{fig:tsne_whole} }
 \subfloat[By our piecewise mappings] { \includegraphics[width=0.48\columnwidth]{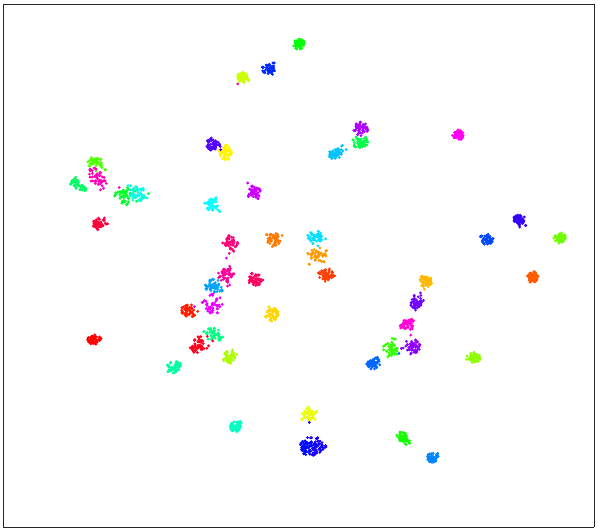} \label{fig:tsne_sep} }
 \caption{Visualization of the category classifiers generated by global mapping and piecewise mappings in 2D space by t-SNE~\cite{tsne}. Each dot denotes a generated classifier and different colors represent different categories. For each category, fifty classifiers are shown, each of which is obtained via randomly sampled five exemplars. This visualization is based on \emph{CUB Birds}. (The figures are best viewed in color.)} \label{fig:tsne}
\end{figure}

As aforementioned, due to high-dimensionality of bilinear feature, it is infeasible to learn a non-linear (even a simple linear) global mapping on the original bilinear features (\emph{e.g.}, $65,536$ dimensionality) in practice. To perform the global mapping, we modify the original AlexNet structure by reducing the number of units of the last convolution layer from $256$ to $64$. By doing this, the bilinear feature becomes $64\times 64=4096$-dimensionality, which is feasible to learn a non-linear global mapping. In experiments, a three-layer MLP acts as the global mapping. The hidden units number is selected via cross-validation based on a set of $\left\{4096, 8192, 16384, 20480\right\}$. Finally, $16,384$ hidden units are selected because of its optimal performance.

For our proposed piecewise mappings, based on the modified BCNN, the piecewise mappings function is applied to $64$-d sub-vectors. Totally, there are $64$ piecewise mappings. Each of them is implemented as a three-layer network whose hidden layers contain $256$ hidden units. ELU~\cite{elu} is used as the activation function for both global mapping and piecewise mappings.

Table~\ref{table:piecewise} demonstrates the comparison results of piecewise mappings \emph{vs.} global mapping. Still the piecewise mappings significantly outperform the global mapping on all the three datasets. These observations can serve as a stronger evidence for the superiority of our proposed method. 

\begin{figure}[t!]
 \centering
 \centering
 \subfloat[Learning curves of the training phrase.] { \includegraphics[width=0.7\columnwidth]{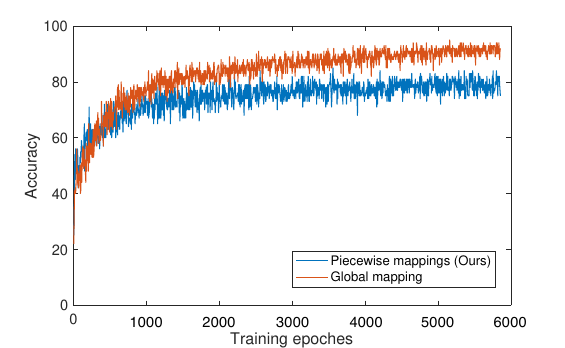} }
 \qquad
 \subfloat[Learning curves of the test phrase.] { \includegraphics[width=0.7\columnwidth]{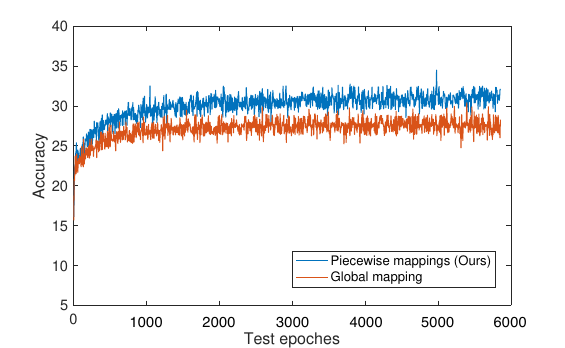} }
 \caption{Comparisons of learning curves of our proposed piecewise mappings with the global mapping. The blue curves indicate the learning behaviors of our proposed piecewise mappings, and the red curves are the global mapping.}
 \label{fig:learningcurves}
\end{figure}

Apart from the above quantitative evaluation, we present some qualitative results by visualizing the $4,096$-d category classifiers generated by global mapping and piecewise mappings in the 2D space in Fig.~\ref{fig:tsne}. The dots with the same color denote the classifiers generated from different exemplar images of the same category in $\mathcal{N}$. Different colors represent classifiers of different categories. We randomly select $250$ exemplars per category to conduct five-shot recognition. Thus, one category contains $50$ versions of classifiers ($50$ dots in the same one color). As shown in the figure, the classifiers generated by piecewise mappings exhibit better category-separability and more centralized intra-category aggregation. This, in some sense, reflects that the classifiers generated by our method tend to capture the essence of the corresponding categories and maintain better distinguishing capacity.

\begin{figure}[t]
 \centering
 \subfloat[1-shot on \emph{CUB Birds}]  { \includegraphics[width=0.48\columnwidth]{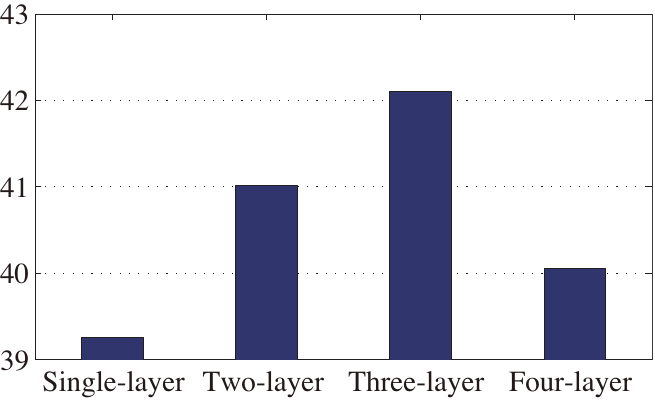} \label{fig:cub1shot} }
 \subfloat[5-shot on \emph{CUB Birds}] { \includegraphics[width=0.48\columnwidth]{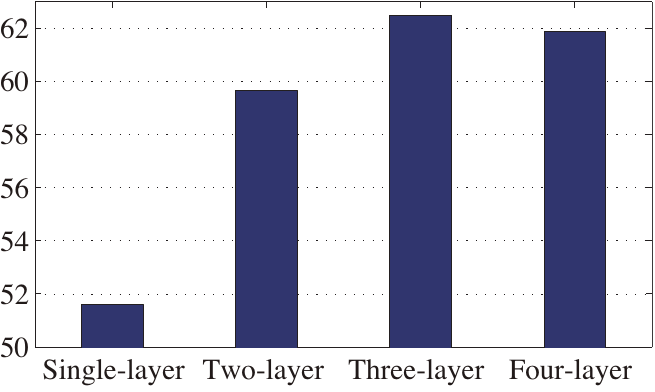} \label{fig:cub5shot} }\\
 \subfloat[1-shot on \emph{Stanford Dogs}]  { \includegraphics[width=0.48\columnwidth]{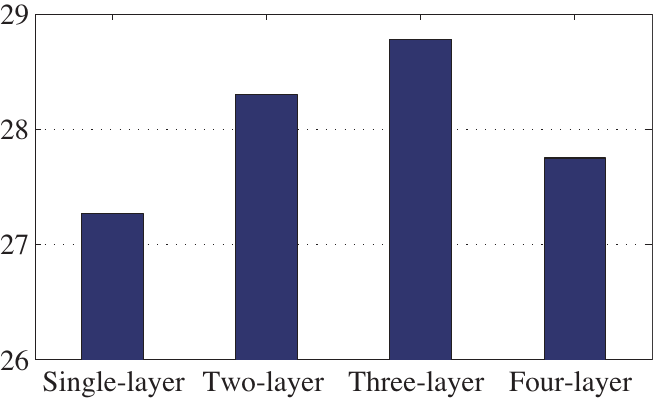} \label{fig:dog1shot} }
 \subfloat[5-shot on \emph{Stanford Dogs}] { \includegraphics[width=0.48\columnwidth]{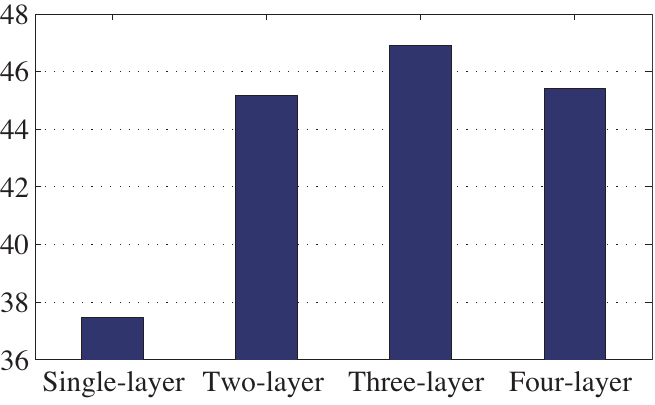} \label{fig:dog5shot} }\\
 \subfloat[1-shot on \emph{Stanford Cars}]  { \includegraphics[width=0.48\columnwidth]{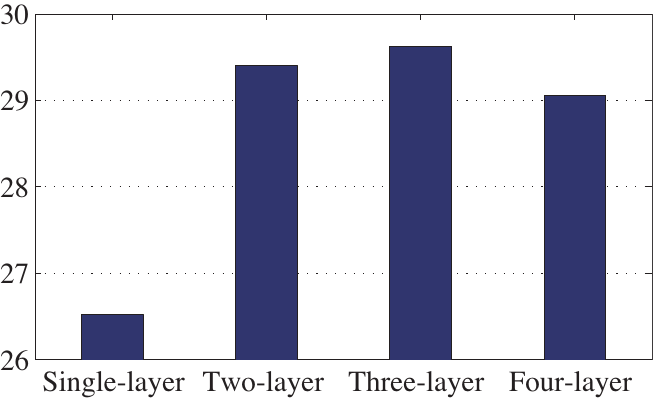} \label{fig:car1shot} }
 \subfloat[5-shot on \emph{Stanford Cars}] { \includegraphics[width=0.48\columnwidth]{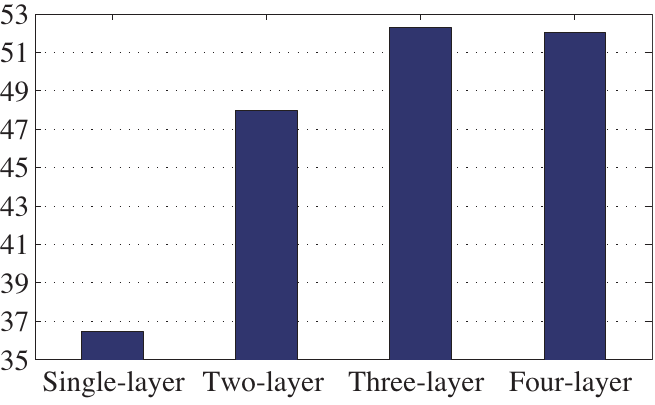} \label{fig:car5shot} }
 \caption{Ablation study on $m_{{\phi}_t}$ with different number of layers. In each sub-figure, the horizontal axis is the number of layers and the vertical axis represents the accuracy rate.} \label{fig:numlayers}
\end{figure}

\begin{figure}[t!]
 \centering
 \subfloat[\emph{CUB Birds}.] { \includegraphics[width=0.5\columnwidth]{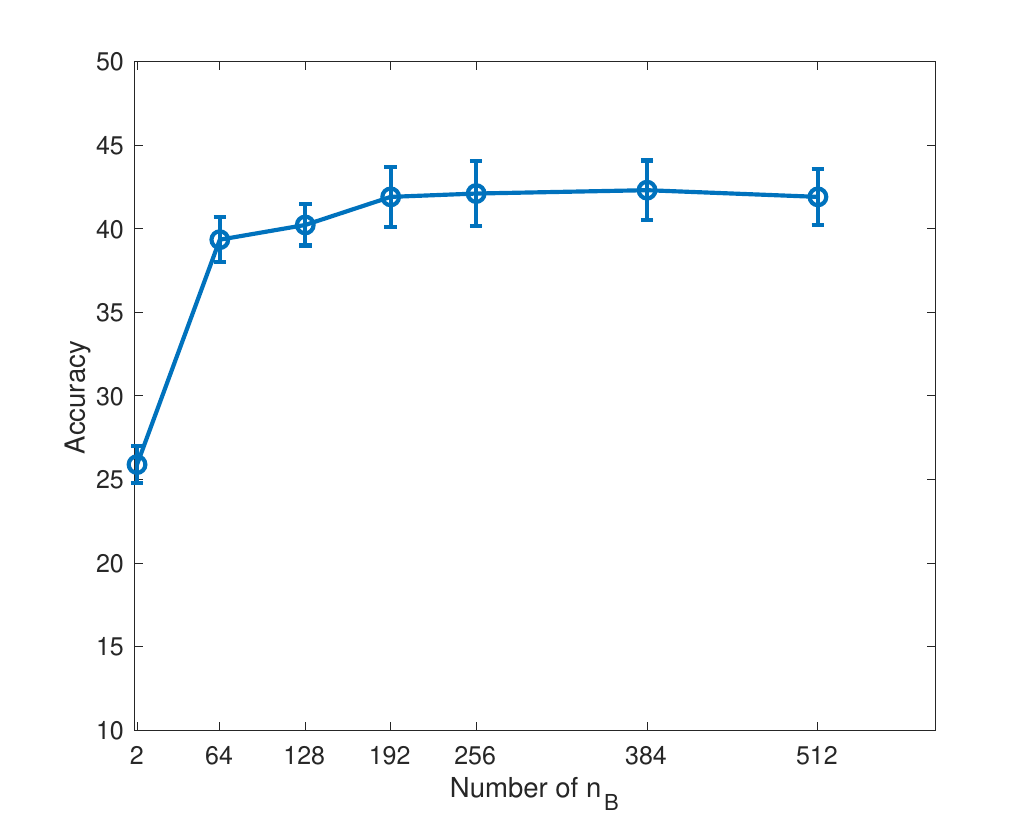} }\\
 \subfloat[\emph{Stanford Dogs}.] { \includegraphics[width=0.5\columnwidth]{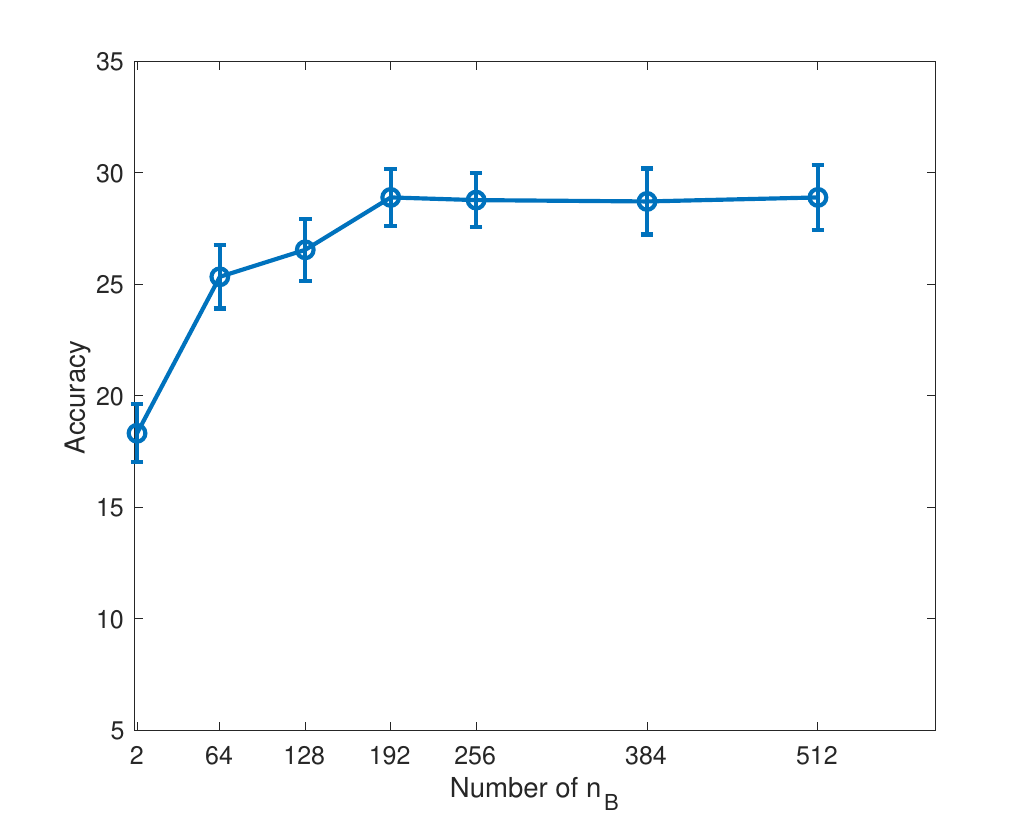} }
 \subfloat[\emph{Stanford Cars}.] { \includegraphics[width=0.5\columnwidth]{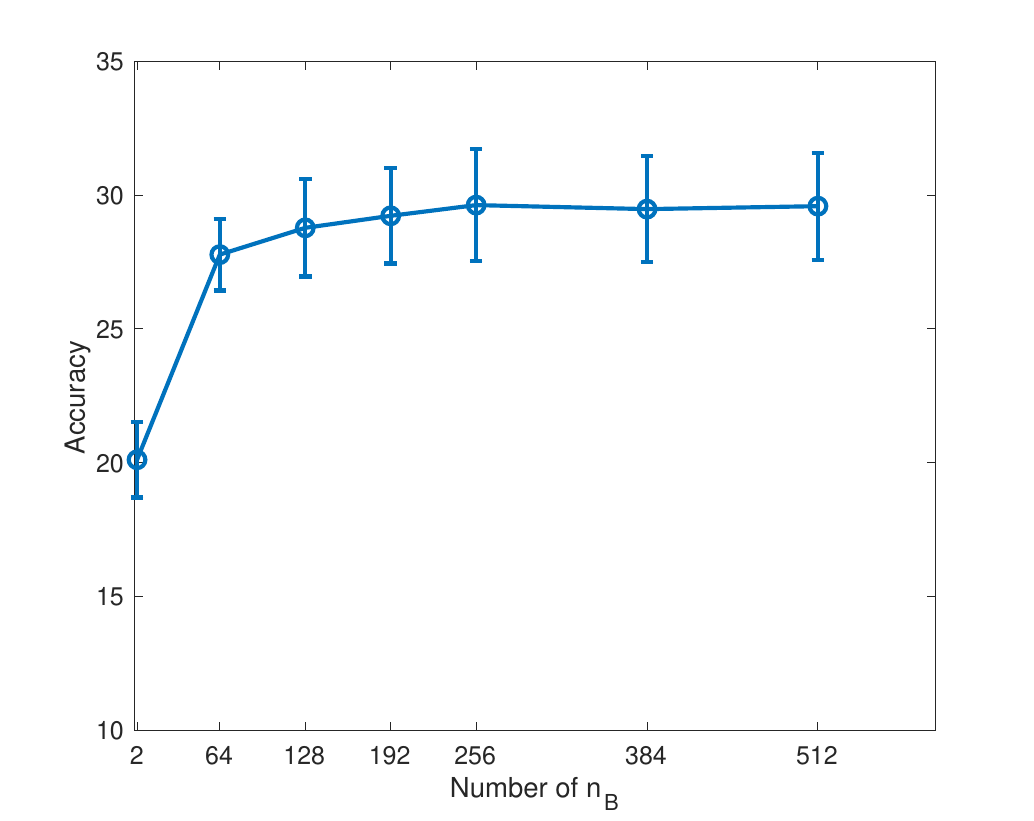} }
 \caption{Comparisons of one-shot accuracy on \emph{CUB Birds}, \emph{Stanford Dogs} and \emph{Stanford Cars} with different numbers of $n_B$.}
 \label{fig:diff_nB}
\end{figure}

On the other hand, in theoretical aspect, the global mapping by fully connected layers is capable to learn the mapping function learned by our piecewise mappings method. In other words, the global mapping should have a larger representation capacity than the proposed piecewise mappings. But, why the global mapping performs worse like above? We hereby show the learning curves of the global mapping and our piecewise mappings in Fig.~\ref{fig:learningcurves}. It is clear to see that the global mapping (\emph{i.e.}, the red curves) achieves higher training accuracy, while it gets worse test accuracy. The observation shows the global mapping has a lower generalization ability, which proves the global mapping is overfitting due to its larger representation capacity. Besides, this looks related to the regularization which constraints the feature mapping happening only in a subset of feature instead of the whole representation. While, thanks to the parameter economy brought by our piecewise mappings, it alleviates overfitting of high dimensional BCNN features.

\subsubsection{$m_{{\phi}_t}$ with different numbers of layers}

We implement the mapping functions $m_{{\phi}_t}$ in our classifier mapping module as MLPs. Since the depth plays an important role in determining the modeling capacity of MLPs, in this part, we investigate how the FSFG performance changes \emph{w.r.t.} different number of layers in $m_{{\phi}_t}$. Specifically, we change the number of layers from $1$ to $4$. The ablation study results are shown in Fig.~\ref{fig:numlayers}.


Generally, we can see that a single-layer mapping leads to worst performance. This is due to its so limited modeling capacity that cannot realize the complex feature-to-classifier mapping. FSFG performance rises when adding another layer and peaks when three-layer mappings are used. Beyond that point, continuing to increase the depth of the mapping functions will do harm to the recognition performance, especially in the one-shot scenario. This study necessitates the need to apply a highly non-linear mapping to learn a satisfactory classifier. 

%

\subsubsection{Different numbers of $n_B$}

In this section, we change the numbers of our piecewise mapping functions as the elements from a set of $\{2, 64, 128, 192, 256, 384, 512\}$. Meanwhile, we fix the number of $n_A$ as $256$. The comparisons are conducted on \emph{CUB Birds}, \emph{Stanford Dogs} and \emph{Stanford Cars} in the one-shot setting. As shown in Fig.~\ref{fig:diff_nB}, it is obvious to observe that, when the number of $n_B$ is large than $128$, there is no significant accuracy gap with the number of $192$, $256$, $384$ and $512$. Except for those, when the number of $n_B$ equals $2$, there is a performance drop due to the extremely small representation ability. The observations of Fig.~\ref{fig:diff_nB} also reveal the stability of our proposed method.

\section{Conclusion}\label{sec:conclude}

In this paper, we have presented the study on fine-grained image recognition in a practical and challenging few-shot learning setting, which requires to learn the classifier for a fine-grained category identified by few exemplars. To address this problem, we proposed an end-to-end trainable network which was inspired by the bilinear CNN model and was tailored for fine-grained few-shot learning. The key novelty of our network was the piecewise classifiers mapping module. By considering the special structure of bilinear CNN features, it decomposed the exemplar-to-classifier mapping into a set of more attainable ``part''-to-``part classifier'' mappings. As a by-product, it significantly reduced the model parameters. Through comprehensive experiments on three popular fine-grained image datasets, our method showed promising results.

In the future, it appears promising to use transfer learning techniques by leveraging the already gained experience (\emph{e.g.}, the classifiers of the known categories) based on the base set for generalizing the learning ability upon the novel set.

%


\bibliographystyle{IEEEtran}
\bibliography{IEEEabrv,FSFG}

\begin{IEEEbiography}[{\includegraphics[width=1in,height=1.25in,clip,keepaspectratio]{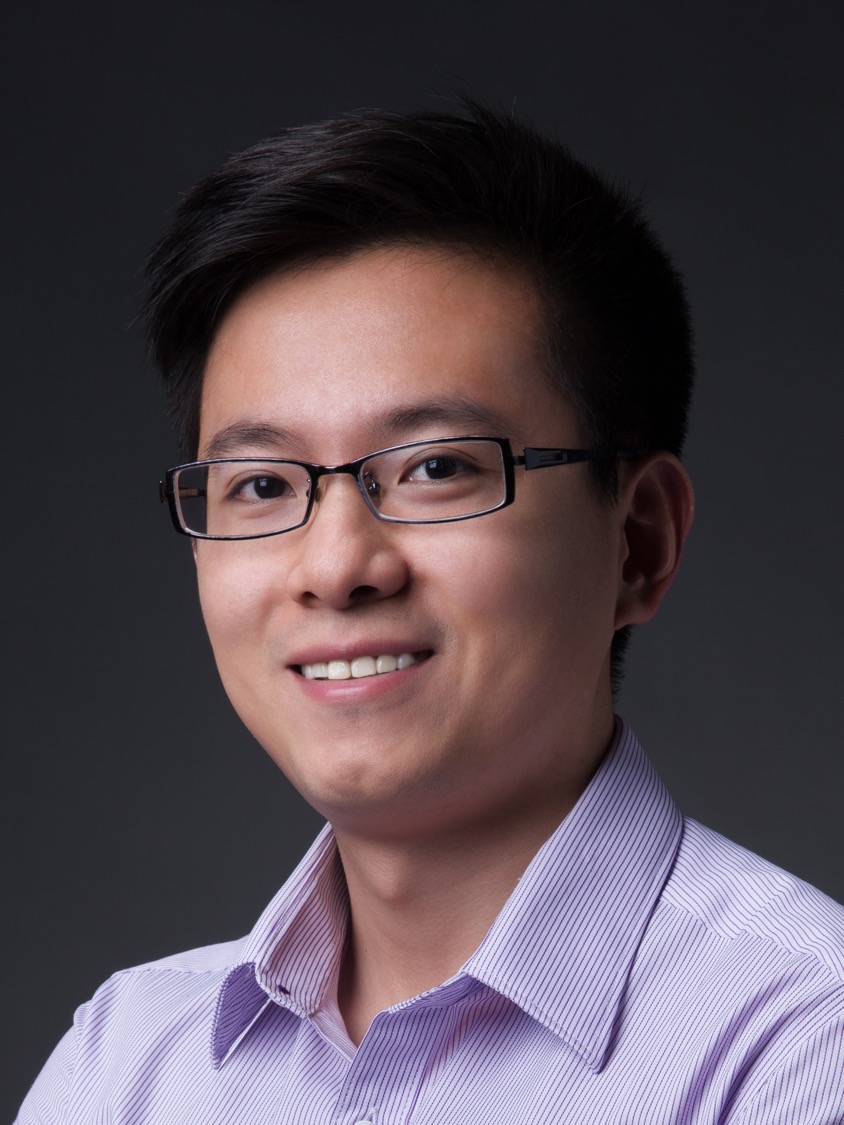}}]{Xiu-Shen Wei}(M'18)
received his BS degree in computer science, and his Ph.D. degree in computer science and technology from Nanjing University. He is now the Research Lead of Megvii Research Nanjing, Megvii Technology, China. He has published about twenty academic papers on the top-tier international journals and conferences, such as IEEE TPAMI, IEEE TIP, IEEE TNNLS, Machine Learning Journal, CVPR, ICCV, IJCAI, etc. He achieved the first place in the Apparent Personality Analysis competition (in association with ECCV 2016), the first place in the iNaturalist competition (in association with CVPR 2019) and the first runner-up in the Cultural Event Recognition competition (in association with ICCV 2015) as the team director. His research interests are computer vision and machine learning. He has served as a PC member of ICCV, CVPR, ECCV, NIPS, IJCAI, AAAI, etc.
\end{IEEEbiography}

\begin{IEEEbiography}[{\includegraphics[width=1in,height=1.25in,clip,keepaspectratio]{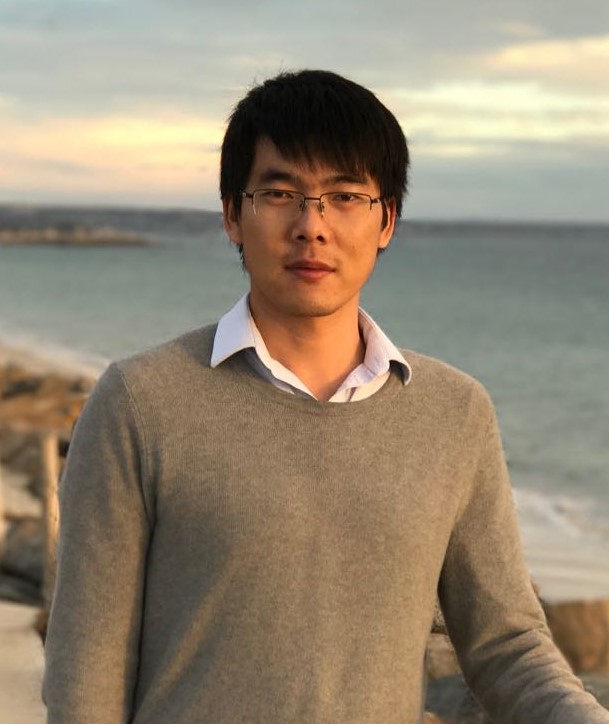}}]{Peng Wang}
received the B.S. and M.S. degrees from School of Electronic and Information Engineering, Beijing Jiaotong University, in 2009 and 2012, respectively, and the Ph.D.degree from the School of Information Technology and Electrical Engineering, The University of Queensland. He is currently a lecturer with University of Wollongong. Prior to this, he was a Post-Doctoral Researcher with The University of Adelaide. His research interests include image classification, video analytics, and deep learning.
\end{IEEEbiography}

\begin{IEEEbiography}[{\includegraphics[width=1in,height=1.25in,clip,keepaspectratio]{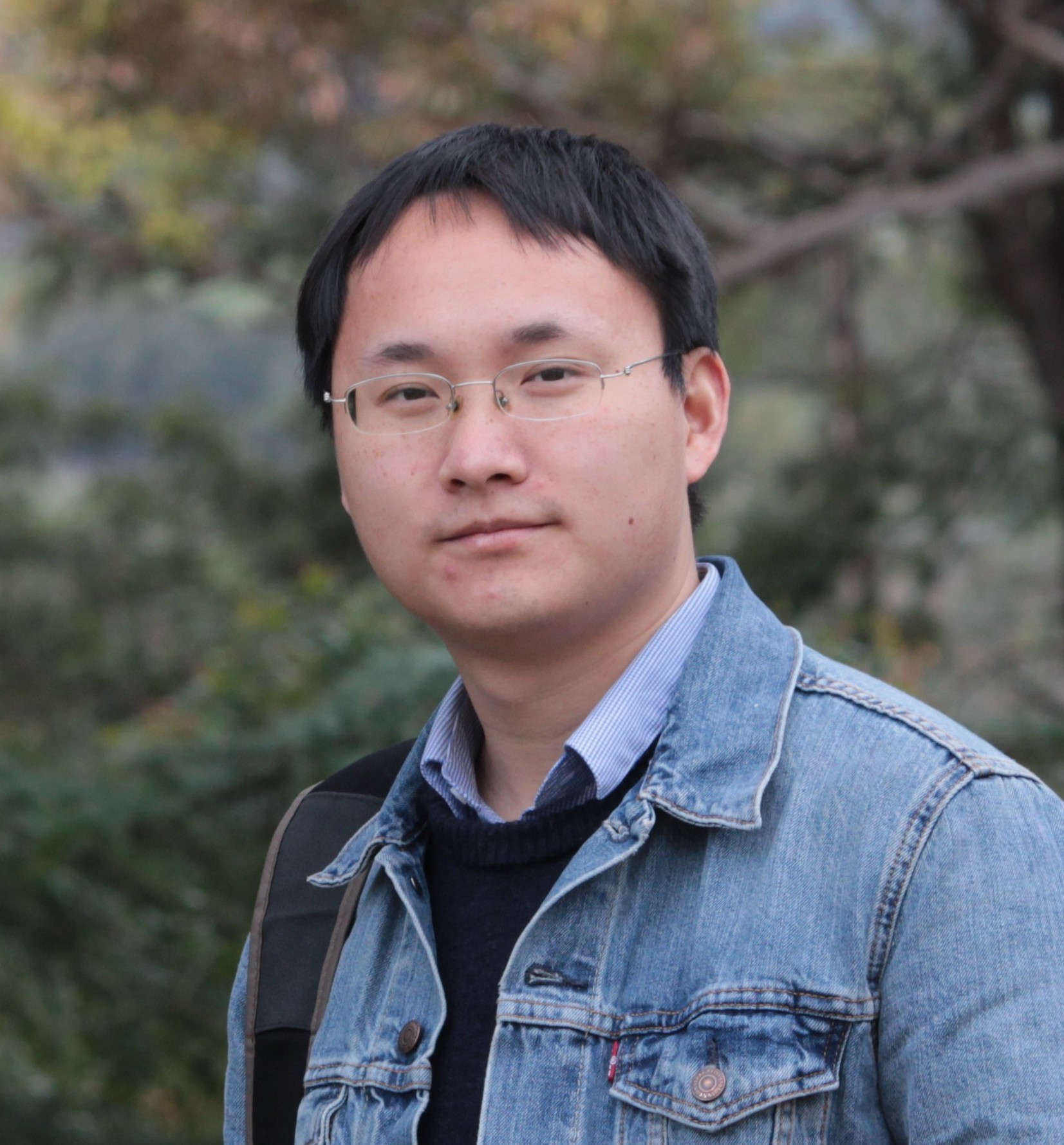}}]{Lingqiao Liu}
received his Ph.D. degree from the Australian National University in 2014. He then joined the University of Adelaide as a research fellow. He is now a DECRA research fellow and Lecturer with the school of computer science at the University of Adelaide. He is the recipient of the Discovery Early Career Researcher Award from the Australia Research Council, and the University of Adelaide Research Fellowship. His current research includes deep learning and its application in computer vision and natural language processing. He served as an associate editor for IEEE Robotics and Automation Letters and a reviewer/PC member for multiple international journals (e.g., TPAMI, TNN, TIP, TCSVT) and conferences (e.g., CVPR, ICCV, IJCAI, AAAI).
\end{IEEEbiography}

\begin{IEEEbiography}[{\includegraphics[width=1in,height=1.25in,clip,keepaspectratio]{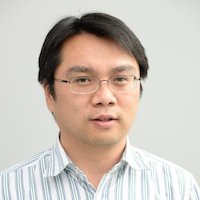}}]{Chunhua Shen}
is a Professor at School of Computer Science, University of Adelaide. He studied at Nanjing University, at Australian National University, and received his PhD degree from the University of Adelaide. From 2012 to 2016, he held an Australian Research Council Future Fellowship.
\end{IEEEbiography}

\begin{IEEEbiography}[{\includegraphics[width=1in,height=1.25in,clip,keepaspectratio]{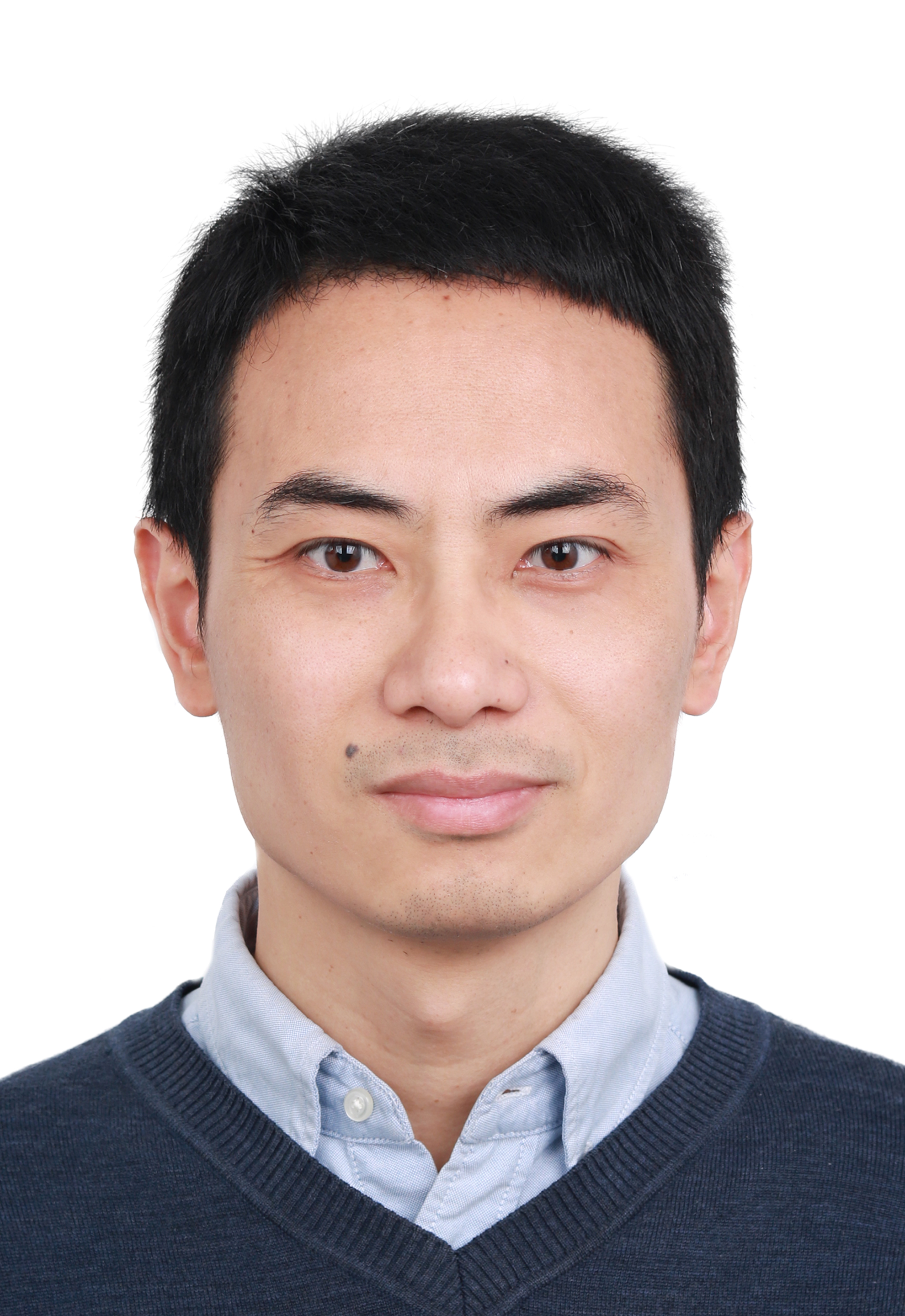}}]{Jianxin Wu}(M'09)
received his BS and MS degrees in computer science from Nanjing University, and his PhD degree in computer science from the Georgia Institute of Technology. He is currently a professor in the Department of Computer Science and Technology at Nanjing University, China, and is associated with the National Key Laboratory for Novel Software Technology, China. He has served as an area chair for CVPR, ICCV and AAAI. His research interests are computer vision and machine learning.
\end{IEEEbiography}

\end{document}